\documentclass[10pt,twocolumn,letterpaper]{article}

\usepackage{iccv}
\usepackage{times}
\usepackage{epsfig}
\usepackage{graphicx}
\usepackage{amsmath}
\usepackage{amssymb}

\usepackage{multirow}
\usepackage{bigstrut}
\usepackage{color}

\usepackage[breaklinks=true,bookmarks=false]{hyperref}

\iccvfinalcopy 

\def\iccvPaperID{10010} 

\ificcvfinal\fi

\begin{document}
\title{Dynamic Implicit Image Function for Efficient Arbitrary-Scale \\Image Representation}

\author{Zongyao He\\
Sun Yat-sen University\\
{\tt\small hezy28@mail2.sysu.edu.cn}
\and Zhi Jin\\
Sun Yat-sen University\\
{\tt\small jinzh26@mail.sysu.edu.cn}
\thanks{This work has been submitted to the IEEE for possible publication. Copyright may be transferred without notice, after which this version may no longer be accessible.}
}

\maketitle
\ificcvfinal\thispagestyle{empty}\fi

\begin{abstract}
Recent years have witnessed the remarkable success of implicit neural representation methods. The recent work Local Implicit Image Function (LIIF) has achieved satisfactory performance for continuous image representation, where pixel values are inferred from a neural network in a continuous spatial domain. However, the computational cost of such implicit arbitrary-scale super-resolution (SR) methods increases rapidly as the scale factor increases, which makes arbitrary-scale SR time-consuming. In this paper, we propose Dynamic Implicit Image Function (DIIF), which is a fast and efficient method to represent images with arbitrary resolution. Instead of taking an image coordinate and the nearest 2D deep features as inputs to predict its pixel value, we propose a coordinate grouping and slicing strategy, which enables the neural network to perform decoding from coordinate slices to pixel value slices. We further propose a Coarse-to-Fine Multilayer Perceptron (C2F-MLP) to perform decoding with dynamic coordinate slicing, where the number of coordinates in each slice varies as the scale factor varies. With dynamic coordinate slicing, DIIF significantly reduces the computational cost when encountering arbitrary-scale SR. Experimental results demonstrate that DIIF can be integrated with implicit arbitrary-scale SR methods and achieves SOTA SR performance with significantly superior computational efficiency, thereby opening a path for real-time arbitrary-scale image representation. Our code can be found at \href{https://github.com/HeZongyao/DIIF}{https://github.com/HeZongyao/DIIF}.
\end{abstract}

\section{Introduction}
Image data has been presented as discrete 2D arrays with fixed resolutions for decades, while our visual world is presented in a continuous and precise manner. Since pixel-based image representation is highly constrained by the resolution, the need to resize images arbitrarily for computer vision tasks has increased rapidly. In addition to traditional interpolation methods, recent Single Image Super-Resolution (SISR) works \cite{lim2017enhanced,zhang2018image,zhang2018residual,chen2021pre,liang2021swinir} have achieved remarkable performance in reconstructing High-Resolution (HR) images from Low-Resolution (LR) ones. However, most of them focus on learning an upsampling function for a specific scale and represent images in a discontinuous manner. Nowadays, arbitrary-scale Super-Resolution (SR) is a raising research topic with great practical significance. 

\begin{figure}[t]
  \centering
  \includegraphics[width=0.87\linewidth]{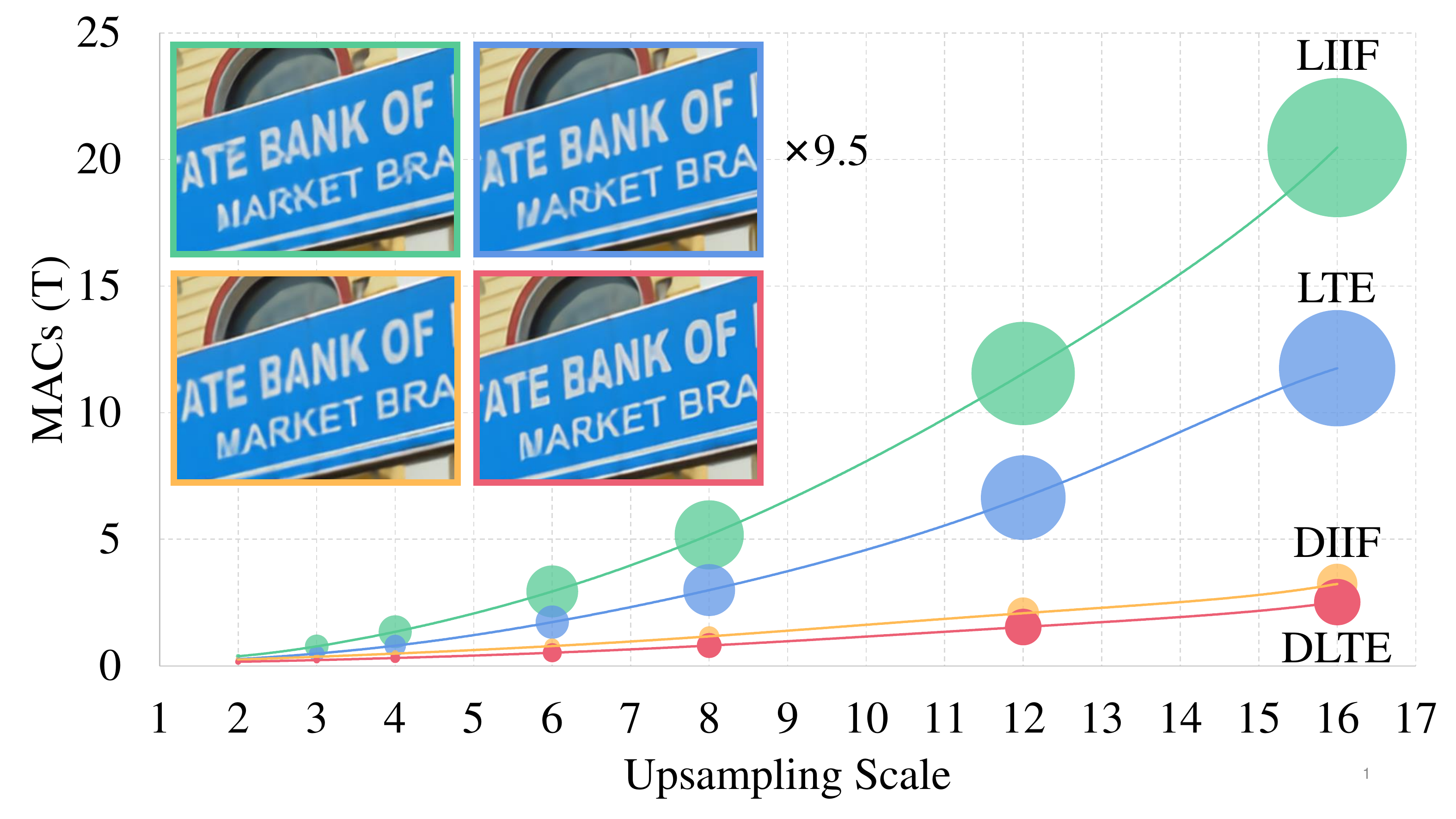}
  \caption{The comparison between LIIF \cite{chen2021learning}, LTE \cite{lee2022local}, and our DIIF and DLTE. The area of the bubbles represents the running time at the scale, and the input is a $320 \times 180$ image for all scales.}
  \label{fig:teaser-figure}
\end{figure}

Neural Radiance Fields (NeRF) and other implicit neural representation methods \cite{chen2019learning,mildenhall2020nerf,park2019deepsdf} have made significant progress in 3D shape reconstruction. The key idea of implicit neural representation is to represent a 3D object or scene as a neural network that maps coordinates to signal values. Inspired by NeRF, the recent work Local Implicit Image Function (LIIF) \cite{chen2021learning} learns to represent 2D images as a continuous implicit function. In LIIF, an image is represented as a set of latent codes generated by an encoder. Given a coordinate and its neighboring latent codes, a decoding Multilayer Perceptron (MLP) predicts the corresponding pixel value at the given coordinate. Since the coordinates are continuous, an image can be presented as LIIF in arbitrary resolution. However, the pixel value at every given coordinate needs to be independently inferred from the decoding function, thereby the computational cost of LIIF increases quadratically as the scale factor increases. Such heavy computational cost makes most image-resizing applications still rely on traditional interpolation methods despite their poor performance, which invisibly creates a huge gap between academic research and practical usage.

To address the aforementioned problem, we propose Dynamic Implicit Image Function (DIIF) for efficient arbitrary-scale image representation, which directly maps coordinate slices to pixel value slices. A coordinate grouping and slicing strategy first aggregates the nearest coordinates around a latent code as a coordinate group, and then divides the coordinate group into slices. The coordinates in the same slice share the same latent code, thereby their pixel values can be predicted by the decoding function simultaneously. Ideally, the computational cost of DIIF depends solely on the input resolution instead of the output resolution. In practice, to maintain the efficiency of DIIF when encountering large-scale SR, dynamic coordinate slicing is proposed to divide a coordinate group into slices of different lengths as the scale factor varies. The vanilla MLP used by LIIF is designed to take a fixed number of coordinates as input, while dynamic coordinate slicing requires the decoder to take a variable number of coordinates as input. Therefore, we propose a Coarse-to-Fine Multilayer Perceptron (C2F-MLP), where the coarse stage network predicts the slice hidden vector based on the input coordinate slice, and the fine stage network independently predicts each pixel value in the given slice using the slice hidden vector.

The idea of using an implicit neural representation for fast and efficient arbitrary-scale SR will provide both precision and convenience to future research and downstream applications. Not only DIIF is an arbitrary-scale image representation that can extrapolate to $\times30$ larger scales, but also the computational cost of DIIF can be $87\%$ lower than LIIF. DIIF can be integrated with 
any implicit arbitrary-scale SR method (\eg LTE \cite{lee2022local}) that inherits the techniques proposed by LIIF. The experimental results demonstrate that DIIF achieves State-Of-The-Art (SOTA) SR performance and computational efficiency on the arbitrary-scale SR task.

Our main contributions are summarized as follows:
\begin{itemize}
\item We propose a novel arbitrary-scale image representation method DIIF, which efficiently predicts the pixel values of coordinate slices by the proposed coordinate grouping and slicing strategy.

\item We propose C2F-MLP to enable efficient decoding with dynamic coordinate slicing, where the number of coordinates in a slice varies as the scale factor varies.

\item DIIF achieves SOTA on arbitrary-scale SR in both efficiency and SR performance. DIIF can be integrated with implicit arbitrary-scale SR methods to significantly reduce their computational cost. 
\end{itemize}

\section{Related Work}
\subsection{Implicit Neural Representation}
Learning implicit neural representation with a parameterized neural network for 3D representation has been an active research topic in recent years. The idea of using an MLP to represent an object has been widely studied in modeling 3D surfaces of the scene \cite{chabra2020deep,jiang2020local,peng2020convolutional,sitzmann2019scene}, the appearance of the 3D structure \cite{Hedman_2021_ICCV,mildenhall2020nerf,niemeyer2020differentiable,Wizadwongsa_2021_CVPR}, and 3D object shapes \cite{atzmon2020sal,chen2019learning,gropp2020implicit,michalkiewicz2019implicit}. Implicit neural representation is capable of capturing the details of the object with a small number of parameters compared with explicit representations, and its differentiable property allows back-propagation through the model for neural rendering \cite{sitzmann2019scene}. Recently, implicit neural representation has made progress in 2D visual tasks, such as image representation \cite{klocek2019hypernetwork,sitzmann2020implicit}, and super-resolution \cite{chen2021learning}. 

\subsection{Single Image Super-Resolution}
SISR as one of the low-level vision tasks has been studied for decades. Dong \etal proposed SRCNN \cite{dong2014learning}, which is a pioneering work of CNN-based SR methods. Later, the field witnessed a variety of network architectures, such as Laplacian pyramid structure \cite{lai2017deep}, residual blocks \cite{ledig2017photo}, densely connected network \cite{tai2017memnet}, and residual-in-residual network \cite{zhang2018image}. Lim \etal proposed EDSR \cite{lim2017enhanced}, which modified SRResNet \cite{ledig2017photo} to construct a more in-depth and broader residual network for better SR performance. Zhang \etal \cite{zhang2018residual} proposed RDN, which boosts SR performance to the next level with Residual Dense Block (RDB). Lately, image processing transformers, such as IPT \cite{chen2021pre} and SwinIR \cite{liang2021swinir}, have been proposed, which surpass CNN-based methods in SR performance using a large dataset. Although previous SISR methods have achieved remarkable performance, most of them are designed for a specific scale.

Arbitrary-scale SR methods are designed to use one network for any scale, which significantly outperform prior works in terms of practice and convenience. Lim \etal proposed MDSR \cite{lim2017enhanced}, which integrates upsampling modules trained for a fixed set of scales, \ie, $\times2$, $\times3$, and $\times4$. Hu \etal proposed MetaSR \cite{hu2019meta}, which generates a convolutional upsampling layer with its meta-network to perform arbitrary-scale SR in its training scales. 
A novel image representation was proposed in LIIF \cite{chen2021learning}, where a pixel value is inferred from a neural network using the corresponding coordinate and neighboring 2D features. Although LIIF achieves robust performance for arbitrary scales up to $\times30$, its computational cost increases rapidly as the scale factor increases. Lee \etal proposed LTE \cite{lee2022local}, a dominant-frequency estimator that enables an implicit function to capture fine details while reconstructing images in a continuous manner. Compared to LIIF and LTE, our DIIF and DLTE representations can achieve SOTA SR performance with significantly less computational cost and running time.

\begin{figure*}[h]
  \centering
  \includegraphics[width=\textwidth]{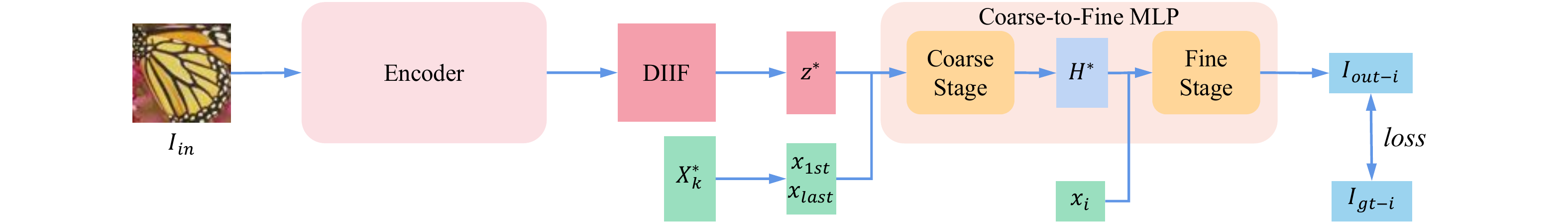}
  \caption{The framework for representing an image as DIIF. An encoder is used to generate the DIIF representation, and the decoder C2F-MLP first takes the boundary coordinates of a coordinate slice $X^{*}_{k}$ and the neighboring latent code $z^{*}$ as inputs to predict the slice hidden vector $H^{*}$, and then predicts the pixel value at any coordinate $x_{i} \in X^{*}_{k}$ using the slice hidden vector.}
  \label{fig:diif-framework}
\end{figure*}

\section{Methods}
The general idea of learning a continuous representation for images is to train an encoder that maps an input image to a 2D feature map as its representation, and a decoding function that maps the query coordinates to the pixel values (\eg, the RGB values) using the neighboring latent codes from the feature map. As shown in Figure \ref{fig:diif-framework}, the encoder maps the input image $I_{in}$ to a 2D feature map $M$ as its DIIF representation. Given the resolution of the ground truth image $I_{gt}$, we can generate a latent code $z^{*}$ from the feature map, and a slice of coordinates $\{x_{1st}, \cdots, x_{last}\}$ around the latent code. The decoding function then takes the generated information to predict the pixel values $\{I_{out}^{1st}, \cdots, I_{out}^{last}\}$ of the coordinate slice. In the training phase, a pixel loss is computed between the predicted pixel values and the pixel values from the ground truth $I_{gt}$. The encoder and the decoding function are jointly trained in a self-supervised SR task, and the learned networks are shared by all images.

\subsection{Dynamic Implicit Image Function}
In DIIF representation, a continuous image $I \in \mathbb{R}^{H \times W \times D_{I}}$ is represented as a 2D feature map $M \in \mathbb{R}^{H \times W \times D_{M}}$ generated by the encoder, and a latent code $z^{*} \in \mathbb{R}^{D_{M}}$ is a feature vector from the feature map $M$. A decoding function $f_{\theta}$ parameterized as a neural network is formulated as:
\begin{equation}
  I(X_{out}) = f_{\theta}(M, X_{out}),
  \label{con:initial-diif}
\end{equation}
where $X_{out}$ are the 2D coordinates of the output pixels in a continuous spatial domain $X \subset \mathbb{R}^{2}$, and $I(X_{out})$ are the corresponding predicted pixel values.

\begin{figure}[t]
  \centering
  \includegraphics[width=\linewidth]{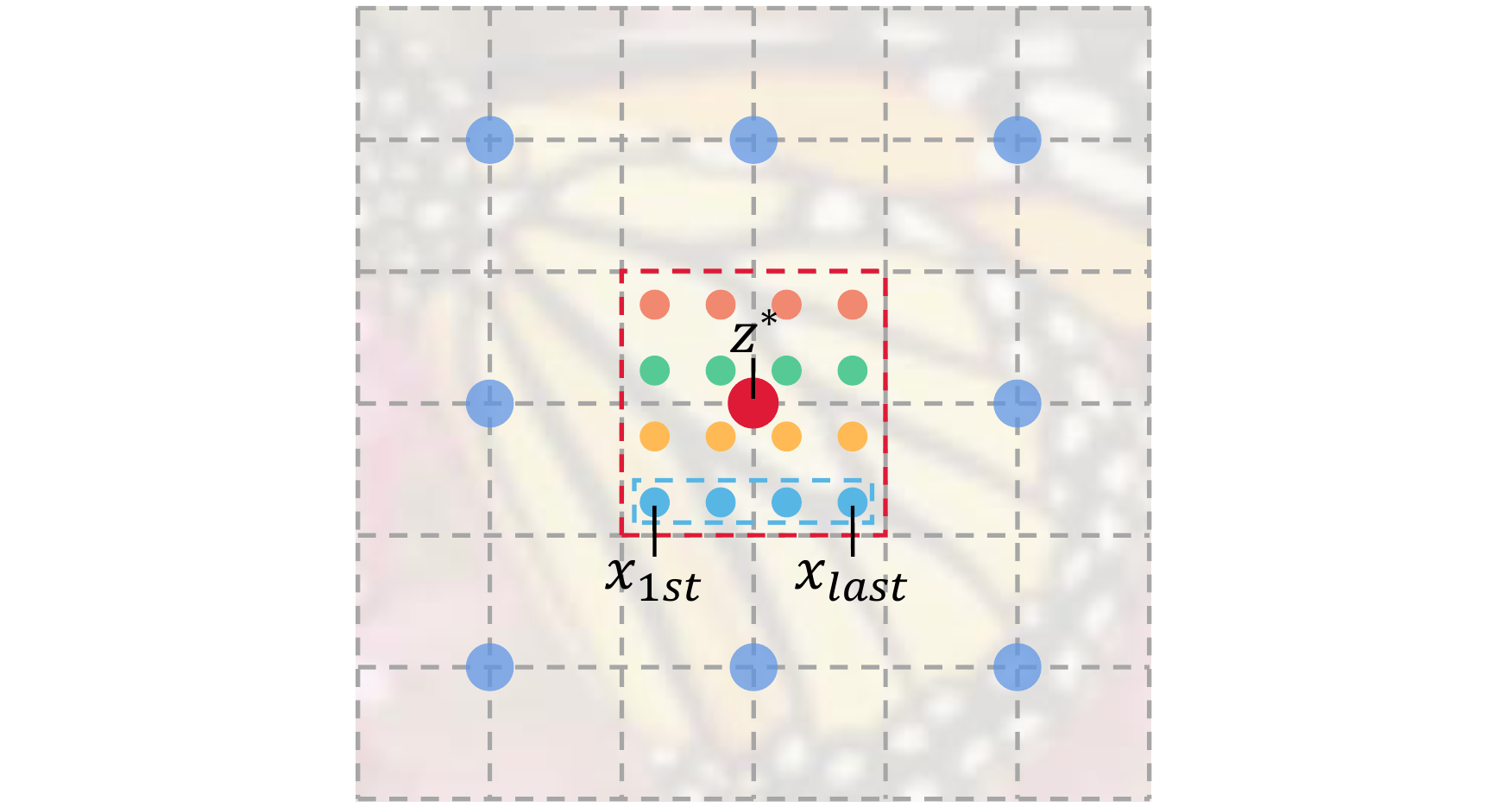}
  \caption{An example of coordinate grouping and slicing with scale $\times 4$ and slice interval $4$. Coordinate grouping aggregates the nearest 16 coordinates around the latent code $z^{*}$ as a coordinate group, and coordinate slicing divides the group into 4 slices.}
  \label{fig:coordinate-grouping-and-slicing}
\end{figure}

\noindent \textbf{Coordinate Grouping and Slicing.}
As shown in Figure \ref{fig:coordinate-grouping-and-slicing}, a coordinate group is defined as a set of coordinates that are closer to the given latent code than other latent codes:
\begin{equation}
  x_{i} \in X^{*}, \Vert x_{i}-v^{*} \Vert_{2} \leq \arg\min_{v \in V}{\Vert x_{i}-v \Vert_{2}},
  \label{con:coordinate-grouping}
\end{equation}
where $X^{*} \subset X$ is the coordinate group corresponding to the latent code $z^{*}$, $v^{*} \in V$ is the coordinate of $z^{*}$, and $V \subset \mathbb{R}^{2}$ is all coordinates of latent codes.

The key idea of coordinate grouping is to share the latent code within a coordinate group so that the decoding function can predict the corresponding pixel values by utilizing the latent code only once. The number of coordinates in one coordinate group is proportional to the scale factor. Therefore, the higher the scale factor is, the more computational cost can be saved. With coordinate grouping, the computational cost of DIIF is determined by the number of coordinate groups, which is equivalent to the input resolution. DIIF with coordinate grouping is formulated as:
\begin{equation}
  I(X^{*}) = f_{\theta}(z^{*}, [x_{tl} - v^{*}, \cdots, x_{br} - v^{*}]),
  \label{con:diif-coordinate-grouping}
\end{equation}
where $X^{*}=\{x_{tl}, x_{tl+1}, \cdots, x_{br}\}$ is the coordinate group, and $x_{tl}$ and $x_{br}$ are the top-left and bottom-right coordinates of the group. The coordinates in a coordinate group are labeled first from left to right, and then from top to bottom.

Coordinate grouping requires the decoding function to predict the pixel values of a coordinate group simultaneously, which imposes a heavy burden on the decoder when encountering large-scale SR. A reasonable solution is to divide a coordinate group into several coordinate slices, and only share the latent code input within one slice instead of the whole group. The number of coordinate slices in a coordinate group is calculated as:
\begin{equation}
  K = Ceil(\frac{g}{u}), 0 < u \leq g,
  \label{con:slice-interval}
\end{equation}
where $Ceil()$ refers to the round-up operation, $K$ is the number of slices, $g$ is the number of coordinates in the given coordinate group, and $u$ is the slice interval that controls the number of coordinates in a slice.

With coordinate grouping and slicing, DIIF is defined as:
\begin{equation}
  \begin{split}
  I(x_{tl}, x_{tl+1}, \cdots, x_{tl+Ku-1})=Concat(\{I(x_{tl+ku},\\
  x_{tl+ku+1}, \cdots, x_{tl+ku+u-1})\}_{k \in \{0, K-1\}}),
  \end{split}
  \label{con:coordinate-slicing}
\end{equation}
where $Concat()$ refers to the concatenation of vectors, and $\{x_{tl}, x_{tl+1}, \cdots, x_{tl+Ku-1}\}$ includes the corresponding coordinate group. In practice, we have to remove the redundant pixel values in the last slice when the number of coordinates in one group is not a multiple of the slice interval.

\begin{figure}[t]
  \centering
  \includegraphics[width=\linewidth]{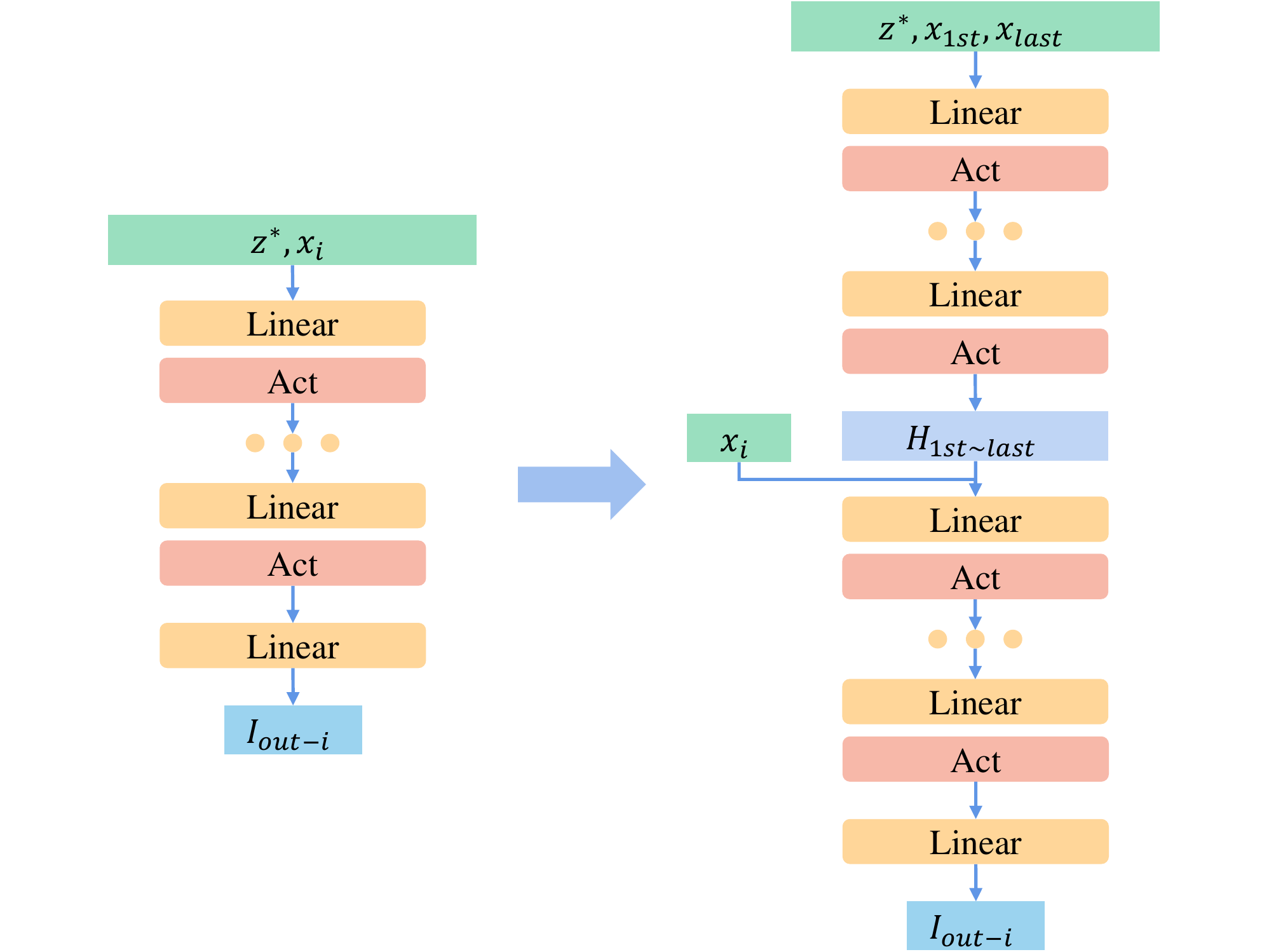}
  \caption{The structure of the vanilla MLP in LIIF \cite{chen2021learning} (left) and our C2F-MLP with dynamic coordinate slicing (right).}
  \label{fig:c2f-mlp}
\end{figure}

\noindent \textbf{Dynamic Coordinate Slicing.}
Coordinate slicing needs to set a suitable slice interval for performance and efficiency balance. The simplest way is fixed coordinate slicing, which sets a fixed slice interval in any case. However, such a strategy retains the quadratic increase of computational cost as the scale factor increases. Moreover, spatial discontinuities and redundant coordinates exist inside the slices. To address these issues, we propose dynamic coordinate slicing to adjust the slice interval as the scale factor varies. The first strategy is linear-order coordinate slicing, which sets the interval as a multiple of the scale factor:
\begin{equation}
  u_{linear} = ns, n \in \mathbb{N}^{+},
  \label{con:linear-order-slicing}
\end{equation}
where $s$ is the scale factor. When this slicing strategy encounters an ideal decoder, the computational cost increases just linearly as the scale factor increases. In practice, the greater the proportion of layers with linear-order slicing applied, the more linear the computational cost increases. We use linear-order slicing with $n=1$ in the final DIIF model.

Another strategy is to set the slice interval to a division of the square of the scale factor with no remainder, which we call constant-order coordinate slicing:
\begin{equation}
  u_{constant} = \frac{s^{2}}{n}, n \in \mathbb{N}^{+},
  \label{con:constant-order-slicing}
\end{equation}
When this slicing strategy encounters an ideal decoder, the computational cost is only determined by the resolution of the input image. In practice, the greater the proportion of layers with constant-order slicing applied, the closer the computational cost is to constant. In the ablation study, we set $n=1$ for DIIF with constant-order slicing. 

\subsection{Coarse-to-Fine MLP}
To perform dynamic coordinate slicing, the decoder needs to have the scalability of taking an indefinite number of coordinates as input and outputting the corresponding pixel values. However, a vanilla MLP only allows fixed-length vectors as input. To address this issue, we propose a C2F-MLP, which divides the decoder into a coarse stage for predicting slice hidden vectors and a fine stage for predicting pixel values. The hidden layer of C2F-MLP consists of a linear layer with the hidden dimension 256, followed by a ReLU. For predicting RGB values, the fine stage uses a linear layer with dimension 3 as the last layer.

\noindent \textbf{MLP Coarse Stage.}
In the coarse stage, the hidden layers parameterized by $\theta_{c}$ take the boundary coordinates of a coordinate slice $X^{*}_{k}$ and its latent code $z^{*}$ as inputs:
\begin{equation}
  H^{*}_{k} = f_{\theta_{c}}(z^{*}, [x_{1st} - v^{*}, x_{last} - v^{*}]),
  \label{con:mlp-coarse-stage}
\end{equation}
where $x_{1st}$ and $x_{last}$ are the first (top-left) and last (bottom-right) coordinates of the slice $X^{*}_{k}$, and $H^{*}_{k}$ is the slice hidden vector corresponding to the slice. 

The slice hidden vector contains the information of the pixel values in the given slice and is used as the fine stage input. The computational cost of the coarse stage is determined by the number of coordinate slices, which is much less than the number of coordinates because of dynamic coordinate slicing. With linear-order slicing, the computational cost of the coarse stage increases linearly as the scale factor increases. The coarse stage also allows the decoding function to utilize the spatial connection within the slice, which makes the pixel value predictions more accurate.

\noindent \textbf{MLP Fine Stage.}
In the fine stage, the hidden layers parameterized by $\theta_{f}$ take the output of the coarse stage, and any coordinate in the given coordinate slice as inputs to predict the pixel value at the coordinate:
\begin{equation}
  I(x_{i}) = f_{\theta_{f}}(H^{*}_{k}, x_{i} - v^{*}),
  \label{con:mlp-fine-stage}
\end{equation}
where $x_{i}$ is a coordinate in the slice $X^{*}_{k}$, and $I(x_{i})$ is the predicted pixel value at $x_{i}$. 

The fine stage is designed to independently predict the pixel value at any coordinate in the given slice. By taking the same structure as the decoder of LIIF, the fine stage ensures that continuous pixel values can be learned by the decoder. The length of slice hidden vectors is 256, which is much shorter than that of latent codes. Therefore, the fine stage uses less computational cost than the decoder of LIIF.

\begin{figure}[t]
  \centering
  \includegraphics[width=\linewidth]{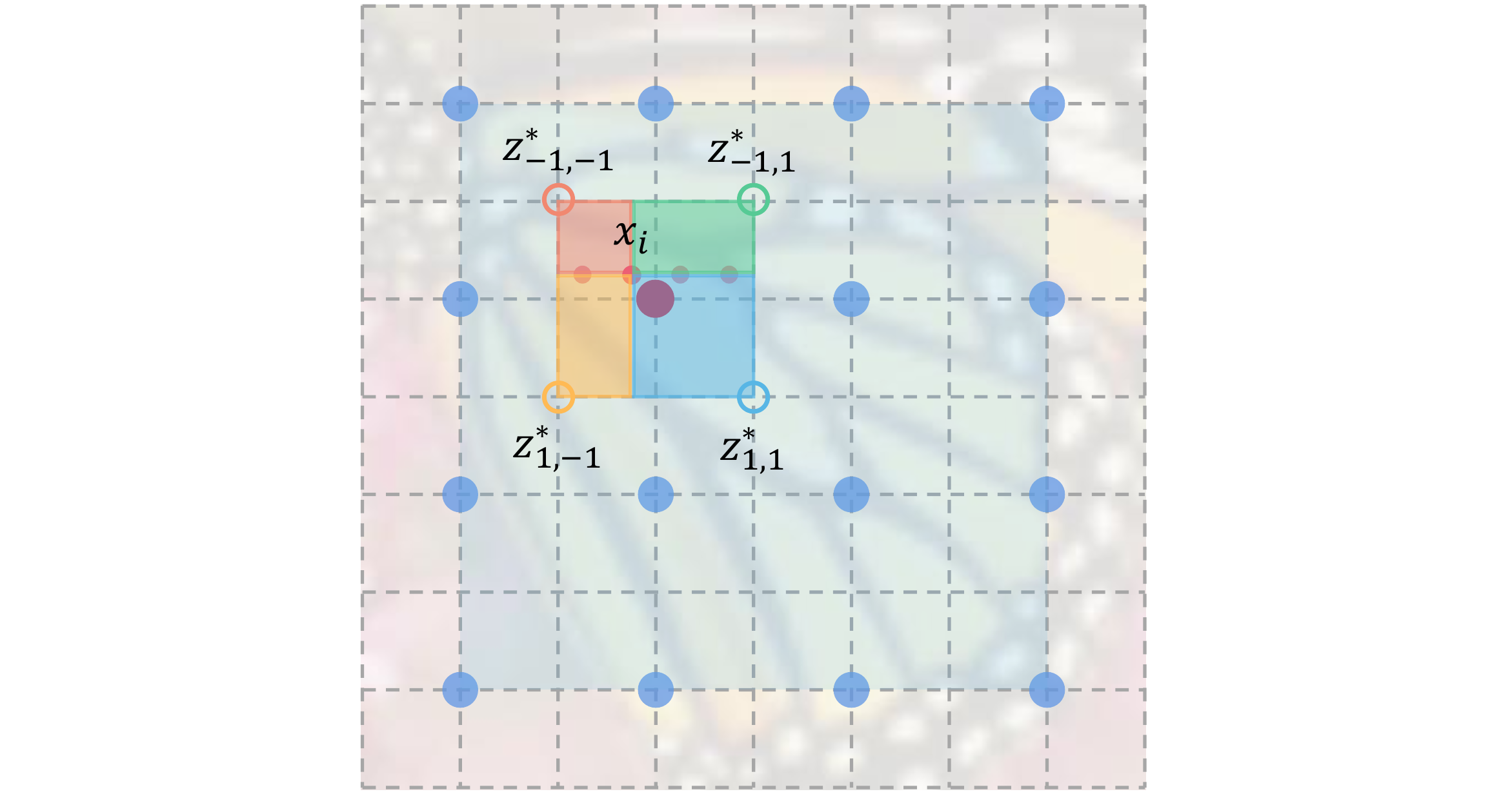}
  \caption{Slice ensemble ensures smooth transitions between coordinate slices. The pixel value at the coordinate $x_{i}$ is predicted by the ensemble of four slice hidden vectors at the vertices of the rectangle $R^{*}$. The bottom-right latent code $z_{1,1}^{*}$ uses the concatenation of its $4 \times 4$ neighboring feature vectors.}
  \label{fig:slice-ensemble}
\end{figure}

\noindent \textbf{Slice Ensemble.}
Using a latent code to predict the nearby pixel values arises the discontinuous prediction issue. As shown in Figure \ref{fig:slice-ensemble}, the coordinates in the coordinate group $X^{*}$ are within the rectangle $R^{*}$ whose vertices are the top-left, top-right, bottom-left, bottom-right coordinates:
\begin{equation}
  X^{*}_{vtx} = \{(-1,-1), (-1,1), (1,-1), (1,1)\},
  \label{con:group-vertices}
\end{equation}

As any coordinates move across the boundary of the rectangle, the coordinate group to which they belong varies. Therefore, the coordinates on different sides of the boundary are predicted from different latent codes, while the distance between them can be infinitely small. To address this issue, instead of using the latent code $z^{*}$ as input to predict the slice hidden vector $H^{*}_{k}$, we extend Eq. \ref{con:mlp-coarse-stage} to:
\begin{equation}
  \begin{split}
  H^{*}_{k, i} = \sum_{t \in X^{*}_{vtx}} \frac{A(v_{t'}^{*},x_{i})}{A_{R^{*}}} H^{*}_{k, t}, \\
  H^{*}_{k, t} = f_{\theta_{c}}(z_{t}^{*}, [x_{1st} - v_{t}^{*}, x_{last} - v_{t}^{*}]),
  \end{split}
  \label{con:slice-ensemble}
\end{equation}
where $x_{i}$ is any query coordinate in the slice $X^{*}_{k}$, $z_{t}^{*}$ is one of the latent codes at the top-left, top-right, bottom-left, bottom-right coordinates $t \in X^{*}_{vtx}$, $v_{t}^{*}$ is the coordinate of $z_{t}^{*}$, $A(v_{t'}^{*},x_{i})$ is the area of the rectangle between the coordinates $x_{i}$ and $v_{t'}^{*}$, $t'$ is diagonal to $t$ (\ie, (1,1) to (-1,-1), (-1,1) to (1,-1)), and $A_{R^{*}}$ is the area of the rectangle $R^{*}$. 

With Eq. \ref{con:slice-ensemble}, the $H^{*}_{k}$ input in Eq. \ref{con:mlp-fine-stage} is replaced by $H^{*}_{k, i}$. It is worth noticing that $H^{*}_{k,t}$ is shared within the coordinate slice so that caching $H^{*}_{k,t}$ can save the computational cost. However, there are no feature vectors at the vertex coordinates. We modify the feature unfolding introduced by LIIF, and use the concatenation of the $4\times4$ neighboring latent codes as the latent code of a vertex coordinate:
\begin{equation}
  z^{*}_{j,k} = Concat(\{z^{*}_{j+l,k+m}\}_{l.m \in \{-1.5,-0.5,0.5,1.5\}}),
  \label{con:feature-unfolding}
\end{equation}
where $z^{*}_{j,k}$ is the latent code at one of the vertex coordinates $(j,k) \in X^{*}_{vtx}$.

We propose slice ensemble by combining Eq. \ref{con:slice-ensemble} and Eq. \ref{con:feature-unfolding}, which first predicts four slice hidden vectors using the vertex latent codes around the center latent code $z^{*}$, and then merges the four predictions by voting with the normalized confidences. The normalized confidence of the slice hidden vector prediction is proportional to the normalized area of the rectangle between the query coordinate and the coordinate of the diagonal latent code. The closer the query coordinate is to the diagonal latent code, the higher the confidence of the prediction is. Slice ensemble not only achieves continuous prediction between adjacent coordinate slices, but also only uses a small amount of computational cost since it is performed in the coarse stage.

\begin{table*}[h]
\centering
\footnotesize
\caption{Quantitative comparison (PSNR (dB)) for arbitrary-scale SR on DIV2K validation set. The efficiency results are evaluated by upsampling a $160\times90$ image to a $2560\times1440$ one. The best results are highlighted in bold. The arbitrary-scale SR methods are trained once for all scales, while the encoder methods train different models for different scales.}
\label{tab:div2k100_psnr_result}
\begin{tabular}{|l|c|c|ccc|ccccc|}
\hline
\multirow{2}{*}{Method} & \multirow{2}{*}{Parameters} & MACs/Running time & \multicolumn{3}{c|}{In-scale} & \multicolumn{5}{c|}{Out-of-scale} \bigstrut[t]\\

& & $\times16$ & $\times2$ & $\times3$ & $\times4$ & $\times6$ & $\times12$ & $\times18$ & $\times24$ & $\times30$ \bigstrut[t]\\
\hline

Bicubic & - & - & 31.01 & 28.22 & 26.66 & 24.82 & 22.27 & 21.00 & 20.19 & 19.59 \bigstrut[t]\\
EDSR-baseline \cite{lim2017enhanced} & 1.813M & 0.200T/0.007s & 34.55 & 30.90 & 28.94 & - & - & - & - & - \bigstrut[t]\\
EDSR-baseline-MetaSR \cite{hu2019meta} & 1.666M & 1.651T/0.653s & 34.64 & 30.93 & 28.92 & 26.61 & 23.55 & 22.03 & 21.06 & 20.37 \bigstrut[t]\\
EDSR-baseline-LIIF \cite{chen2021learning} & 1.567M & 5.117T/3.923s & 34.67 & 30.96 & 29.00 & 26.75 & 23.71 & 22.17 & 21.18 & 20.48 \bigstrut[t]\\
EDSR-baseline-DIIF (ours) & 1.683M & 0.790T/\textbf{0.336s} & 34.69 & 31.00 & 29.03 & 26.79 & 23.75 & 22.21 & 21.22 & 20.52 \bigstrut[t]\\
EDSR-baseline-LTE \cite{lee2022local} & 1.714M & 2.936T/1.728s & 34.72 & 31.02 & 29.04 & 26.81 & 23.78 & \textbf{22.23} & \textbf{21.24} & \textbf{20.53} \bigstrut[t]\\
EDSR-baseline-DLTE (ours) & \textbf{1.559M} & \textbf{0.613T}/0.428s & \textbf{34.74} & \textbf{31.04} & \textbf{29.07} & \textbf{26.82} & \textbf{23.79} & \textbf{22.23} & \textbf{21.24} & \textbf{20.53}\bigstrut[t]\\

\hline

RDN \cite{zhang2018residual} & 22.567M & 0.492T/0.031s & 35.05 & 31.28 & 29.28 & - & - & - & - & - \bigstrut[t]\\
RDN-MetaSR \cite{hu2019meta} & 22.419M & 1.950T/0.704s & 35.00 & 31.27 & 29.25 & 26.88 & 23.73 & 22.18 & 21.17 & 20.47 \bigstrut[t]\\
RDN-LIIF \cite{chen2021learning} & 22.321M & 5.416T/3.979s & 34.99 & 31.26 & 29.27 & 26.99 & 23.89 & 22.34 & 21.31 & 20.59 \bigstrut[t]\\
RDN-DIIF (ours) & 22.437M & 1.108T/\textbf{0.384s} & 35.02 & 31.31 & 29.32 & 27.04 & 23.95 & 22.39 & 21.36 & 20.64 \bigstrut[t]\\
RDN-LTE \cite{lee2022local} & 22.468M & 3.235T/1.784s & 35.04 & 31.32 & 29.33 & 27.04 & 23.95 & 22.40 & 21.36 & 20.64 \bigstrut[t]\\
RDN-DLTE (ours) & \textbf{22.313M} & \textbf{0.905T}/0.475s & \textbf{35.11} & \textbf{31.39} & \textbf{29.39} & \textbf{27.10} & \textbf{24.01} & \textbf{22.42} & \textbf{21.39} & \textbf{20.66}\bigstrut[t]\\

\hline


\end{tabular}
\end{table*}

\begin{table*}[h]
\centering
\footnotesize
\caption{Quantitative comparison (PSNR (dB)) for arbitrary-scale SR on benchmark test sets. The best results are highlighted in bold. All of the arbitrary-scale SR methods use RDN as the encoder.}
\label{tab:benchmark_result}
\begin{tabular}{|l|ccc|cc|ccc|cc|ccc|cc|}
\hline
\multirow{3}{*}{Method} & \multicolumn{5}{c|}{Set14} & \multicolumn{5}{c|}{BSD100} & \multicolumn{5}{c|}{Urban100}\bigstrut[t]\\
\cline{2-16}
& \multicolumn{3}{c|}{In-scale} & \multicolumn{2}{c|}{Out-of-scale}
& \multicolumn{3}{c|}{In-scale} & \multicolumn{2}{c|}{Out-of-scale} 
& \multicolumn{3}{c|}{In-scale} & \multicolumn{2}{c|}{Out-of-scale}\bigstrut[t]\\

& $\times2$ & $\times3$ & $\times4$ & $\times6$ & $\times8$ 
& $\times2$ & $\times3$ & $\times4$ & $\times6$ & $\times8$ 
& $\times2$ & $\times3$ & $\times4$ & $\times6$ & $\times8$ \bigstrut[t]\\
\hline

RDN \cite{zhang2018residual} 
& 34.01 & 30.57 & 28.81 & - & - 
& 32.34 & 29.26 & 27.72 & - & - 
& 32.89 & 28.80 & 26.61 & - & -
\bigstrut[t]\\

MetaSR \cite{hu2019meta} 
& 33.98 & 30.54 & 28.78 & 26.51 & 24.97 
& 32.33 & 29.26 & 27.71 & 25.90 & 24.83 
& 32.92 & 28.82 & 26.55 & 23.99 & 22.59 
\bigstrut[t]\\

LIIF \cite{chen2021learning} 
& 33.97 & 30.53 & 28.80 & 26.64 & 25.15 
& 32.32 & 29.26 & 27.74 & 25.98 & 24.91 
& 32.87 & 28.82 & 26.68 & 24.20 & 22.79 
\bigstrut[t]\\

DIIF (ours)
& 34.00 & 30.61 & 28.86 & 26.69 & 25.19 
& 32.27 & 29.28 & 27.75 & 25.98 & 24.94 
& 32.97 & 28.90 & 26.79 & 24.23 & 22.84 
\bigstrut[t]\\

LTE \cite{lee2022local} 
& \textbf{34.09} & 30.58 & 28.88 & 26.71 & 25.16 
& \textbf{32.36} & 29.30 & 27.77 & 26.01 & 24.95
& 33.04 & 28.97 & 26.81 & 24.28 & 22.88 
\bigstrut[t]\\

DLTE (ours)
& 33.99 & \textbf{30.62} & \textbf{28.93} & \textbf{26.75} & \textbf{25.25} 
& 32.27 & \textbf{29.33} & \textbf{27.79} & \textbf{26.02} & \textbf{24.97} 
& \textbf{33.13} & \textbf{29.02} & \textbf{26.90} & \textbf{24.35} & \textbf{22.95} 
\bigstrut[t]\\

\hline
\end{tabular}
\end{table*}

\section{Experiments}
\subsection{Implementation and Evaluation Settings}
Similar to LIIF, we use the first 800 images from DIV2K dataset \cite{agustsson2017ntire} as the training set and another 100 images for testing. The goal of DIIF is to generate continuous representations for pixel-based images. Therefore, we uniformly sample the downsampling scale from the range of $[\times2, \times4]$ by $0.5$ interval in each training iteration, and use the sampled scale to downsample the HR images with bicubic interpolation. The spatial size of the cropped HR patch is $192 \times 192$. 
For data argumentation, we utilize random horizontal flips and 90-degree rotations on the training images. 

We use a 5-layer C2F-MLP as the decoder, where the coarse stage consists of 2 hidden layers, and the fine stage consists of 2 hidden layers and 1 linear layer for output. The previous arbitrary-scale SR methods are often combined with different encoders to demonstrate their effectiveness. We follow LIIF to use EDSR-baseline \cite{lim2017enhanced} and RDN \cite{zhang2018residual} 
, respectively, as the encoder. We train our models for $1 \times 10^{6}$ iterations (\ie, $1000$ epochs) with batch size 16 and use L1 pixel loss as the loss function. We use an initial learning rate of $1 \times 10^{-4}$ and decay the learning rate by a factor of $0.5$ every $2 \times 10^{5}$ iterations. For optimization, we use Adam \cite{kingma2015adam} with $\beta_{1} = 0.9$, $\beta_{2} = 0.999$. We implement our models with the PyTorch framework. 

We evaluate the models on DIV2K validation set and three public benchmark test sets – Set14 \cite{zeyde2010single}, BSD100 \cite{martin2001database}, and Urban100 \cite{huang2015single}. 
Similar to LIIF and LTE, we use PSNR for performance evaluation. For efficiency evaluation, we use Multiply–Accumulate Operations (MACs) to evaluate the computational cost and use running time on an RTX 2080 Ti GPU to evaluate the model speed. We also provide the number of parameters for comprehensive evaluation. 

\begin{figure*}[h]
  \centering
  \includegraphics[width=0.90\textwidth]{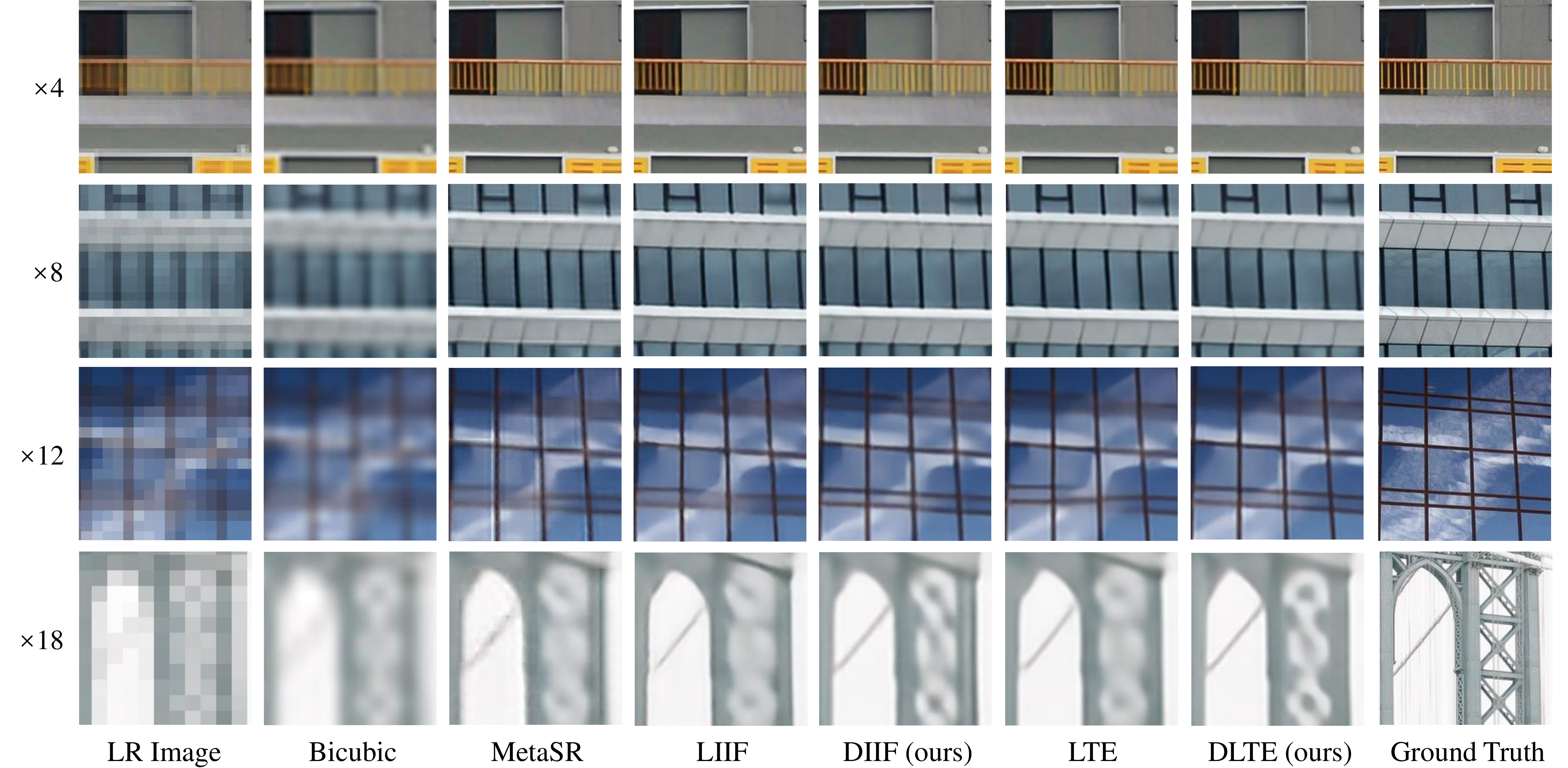}
  \caption{Qualitative comparison for arbitrary-scale SR. All of the arbitrary-scale SR methods use RDN as the encoder.}
  \label{fig:qualitative_comparison}
\end{figure*}

\begin{table*}[h]
\centering
\footnotesize
\caption{The computational cost (MACs) and running time (GPU-seconds) comparison for arbitrary-scale SR. The best results are highlighted in bold. All of the methods use EDSR-baseline as the encoder. The resolution of the input is fixed at $320 \times 180$.}
\label{tab:computation_comparison}
\begin{tabular}{|l|c|ccc|cccc|}
\hline
\multirow{2}{*}{Method} & Parameters & \multicolumn{3}{c|}{In-scale} & \multicolumn{4}{c|}{Out-of-scale} \bigstrut[t]\\

& w/o encoder & $\times2$ & $\times3$ & $\times4$ & $\times6$ & $\times12$ & $\times18$ & $\times24$ \bigstrut[t]\\ 
\hline

MetaSR \cite{hu2019meta} & 457K & 0.17T/\textbf{0.02s} & 0.29T/0.06s & 0.47T/0.16s & 0.97T/0.34s & 3.75T/1.47s & 8.34T/3.39s & 14.77T/6.05s  \bigstrut[t]\\ 

LIIF \cite{chen2021learning} & 355K & 0.38T/\textbf{0.02s} & 0.77T/0.51s & 1.35T/1.00s & 2.94T/2.47s & 11.55T/9.81s & 25.89T/22.06s & 45.97T/39.20s \bigstrut[t]\\ 

DIIF (ours) & 474K & 0.28T/0.03s & 0.36T/\textbf{0.03s} & 0.48T/\textbf{0.04s} & 0.78T/\textbf{0.22s} & 2.08T/\textbf{0.93s} & 3.91T/\textbf{2.01s} & 6.28T/\textbf{3.43s} \bigstrut[t]\\ 

LTE \cite{lee2022local} & 506K & 0.26T/\textbf{0.02s} & 0.49T/0.22s & 0.80T/0.43s & 1.73T/1.02s & 6.64T/6.60s & 14.84T/16.79s & 26.32T/30.97s \bigstrut[t]\\ 

DLTE (ours) & \textbf{347K} & \textbf{0.16T}/0.03s & \textbf{0.23T}/0.04s & \textbf{0.31T}/0.09s & \textbf{0.52T}/0.34s & \textbf{1.54T}/1.23s & \textbf{3.09T}/2.43s & \textbf{5.19T}/4.14s \bigstrut[t]\\ 
\hline
\end{tabular}
\end{table*}

\subsection{Arbitrary-Scale SR Comparison}
\noindent \textbf{Quantitative Results.}
In Table \ref{tab:div2k100_psnr_result}, we compare our DIIF and DLTE methods with the reference arbitrary-scale SR methods on DIV2K validation set. We also provide the results of bicubic interpolation and the encoder methods with their initial upsampling modules. Following LIIF, we train different encoder models for in-distribution scales $\times 2, \times 3, \times 4$. For efficiency evaluation, we provide the computational cost (MACs) and running time results on the $\times16$ SR task, upsampling a $160\times90$ image to a $2560\times1440$ one. 

The PSNR results in Table \ref{tab:div2k100_psnr_result} confirm that with the same encoder, our DIIF and DLTE consistently achieve better SR performance for all scales compared to LIIF and LTE, respectively. Specifically, DLTE achieves SOTA SR performance with minimal computational cost by integrating DIIF with the performance boosters proposed by LTE. Although DIIF is a representation designed for efficiency improvement, it can enhance the SR performance of LIIF and LTE.
Instead of using the manually defined latent code for each pixel as LIIF, DIIF generates a slice hidden vector in the coarse stage to utilize the spatial connection within a coordinate slice. The slice hidden vector enables a shallow fine stage to predict precise and continuous pixel values. As reported in Table \ref{tab:ablation_study}, the comparison between DIIF with 5 layers in the fine stage (the last row) and LIIF confirms that the learned slice hidden vector is superior to the manually defined latent code in representing images.

The MACs results in Table \ref{tab:div2k100_psnr_result} indicate that DIIF and DLTE use significantly less computational cost than MetaSR, LIIF, and LTE. With superior PSNR performance, DIIF reduces $85\%$ of the computational cost of LIIF, and DLTE reduces $79\%$ of the computational cost of LTE when the scale is $\times16$. The running time results in Table \ref{tab:div2k100_psnr_result} also confirm that DIIF and DLTE are significantly faster than LIIF and LTE. A minor defect of DIIF is that it has limited effectiveness in reducing parameters. However, most of the parameters come from the encoder, thereby the differences in the parameters of the decoder are negligible. 

In Table \ref{tab:benchmark_result}, we report the PSNR comparison on the benchmark test sets for comprehensive evaluation. All of the arbitrary-scale SR methods are trained once for all scales and use RDN as the encoder. The PSNR results verify that DIIF and DLTE outperform LIIF and LTE for most scales, respectively. Overall, the results in Table \ref{tab:div2k100_psnr_result} and Table \ref{tab:benchmark_result} confirm that the effectiveness of DIIF in precision and efficiency is universally applicable to different encoders, implicit arbitrary-scale SR methods, test sets, and scales. 


\noindent \textbf{Qualitative Results.}

We show the qualitative comparison for arbitrary-scale SR in Figure \ref{fig:qualitative_comparison}. All of the methods use RDN as the encoder. As shown in the figure, MetaSR, LIIF, and LIE generate blurry results or results with structural distortions. MetaSR further shows discontinuity and limited precision in large-scale results. Compared to LIIF and LTE, DIIF and DLTE are capable of providing visually pleasing results with sharper edges and fewer structural distortions for both in-distribution and out-of-distribution scales.

\noindent \textbf{Efficiency Results.}
In Table \ref{tab:computation_comparison}, we compare the computational cost (MACs), running time, and parameters of arbitrary-scale SR methods on multiple scales. All of the methods use EDSR-baseline as the encoder. The resolution of the input image is fixed at $320 \times 180$. The running time is evaluated on an RTX 2080 Ti GPU and each result is the average of 100 runs. The MACs results demonstrate that DIIF and DLTE save at least $70\%$ of computational cost for out-of-distribution scales compared with LIIF and LTE, respectively. Since the computational cost of the coarse stage increases linearly as the scale factor increases, the larger the scale factor is, the greater the proportion of computational cost that can be saved by our methods. The running time results further confirm that DIIF is more than 10 times faster than LIIF, and DLTE is more than 5 times faster than LTE from scale $\times3$ to $\times24$. As the scale factor increases, the running time of DIIF increases slower than the reference methods, which brings convenience to practical application.

\begin{table*}[h]
\centering
\footnotesize
\caption{Ablation study on design choices of DIIF. The PSNR (dB) results are evaluated on DIV2K validation set, and the MACs results are evaluated by inputting $160\times90$ images. The best results are highlighted in bold. All of the models use EDSR-baseline as the encoder. ds/se/cs refer to dynamic coordinate slicing, slice ensemble, and constant-order coordinate slicing, respectively. 
}
\label{tab:ablation_study}
\begin{tabular}{|l|c|c|c|ccc|ccccc|}
\hline
\multirow{2}{*}{Method} & \multirow{2}{*}{Decoder layers} & Parameters & MACs & \multicolumn{3}{c|}{In-scale} & \multicolumn{5}{c|}{Out-of-scale} \bigstrut[t]\\

& & w/o encoder & $\times16$ & $\times2$ & $\times3$ & $\times4$ & $\times6$ & $\times12$ & $\times18$ & $\times24$ & $\times30$\bigstrut[t]\\
\hline

LIIF & $[4\times256, 3]$ & \textbf{355K} & 5.117T & 34.67 & 30.96 & 29.00 & 26.75 & 23.71 & 22.17 & 21.18 & 20.48 \bigstrut[t]\\

DIIF (w/o ds) & $[4\times256, 3]$  & 359K & 1.328T & 34.68 & 30.97 & 29.01 & 25.14 & 22.94 & 20.86 & 20.51 & 19.64 \bigstrut[t]\\
DIIF (w/o se) & $[2\times256]+[2\times256, 3]$  & 356K & 0.542T & 34.68 & 30.99 & 29.02 & 26.77 & 23.75 & 22.20 & 21.21 & 20.51 \bigstrut[t]\\
DIIF ($\times 4$) & $[2\times256]+[2\times256, 3]$  & 474K & 0.790T & 33.70 & 30.85 & 29.01 & 26.78 & 23.74 & 22.18 & 21.21 & 20.52 \bigstrut[t]\\
DIIF (cs) & $[2\times256]+[2\times256, 3]$  & 474K & \textbf{0.512T} & \textbf{34.69} & \textbf{31.01} & \textbf{29.04} & 26.79 & 23.75 & \textbf{22.22} & 21.22 & 20.52 \bigstrut[t]\\
\hline

DIIF & $[2\times256]+[2\times256, 3]$ & 474K & 0.790T & \textbf{34.69} & 31.00 & 29.03 & 26.79 & 23.75 & 22.21 & 21.22 & 20.52 \bigstrut[t]\\
DIIF (c4+f3)  & $[4\times256]+[2\times256, 3]$ & 609K & 0.908T & 34.68 & 31.00 & \textbf{29.04} & 26.79 & 23.75 & 22.21 & 21.22 & 20.52 \bigstrut[t]\\
DIIF (c2+f5)  & $[2\times256]+[4\times256, 3]$ & 609K & 1.292T & \textbf{34.69} & \textbf{31.01} & \textbf{29.04} & \textbf{26.80} & \textbf{23.76} & \textbf{22.22} & \textbf{21.23} & \textbf{20.53} \bigstrut[t]\\
\hline
\end{tabular}
\end{table*}

\subsection{Ablation Study}
\begin{figure}[t]
  \centering
  \includegraphics[width=\linewidth]{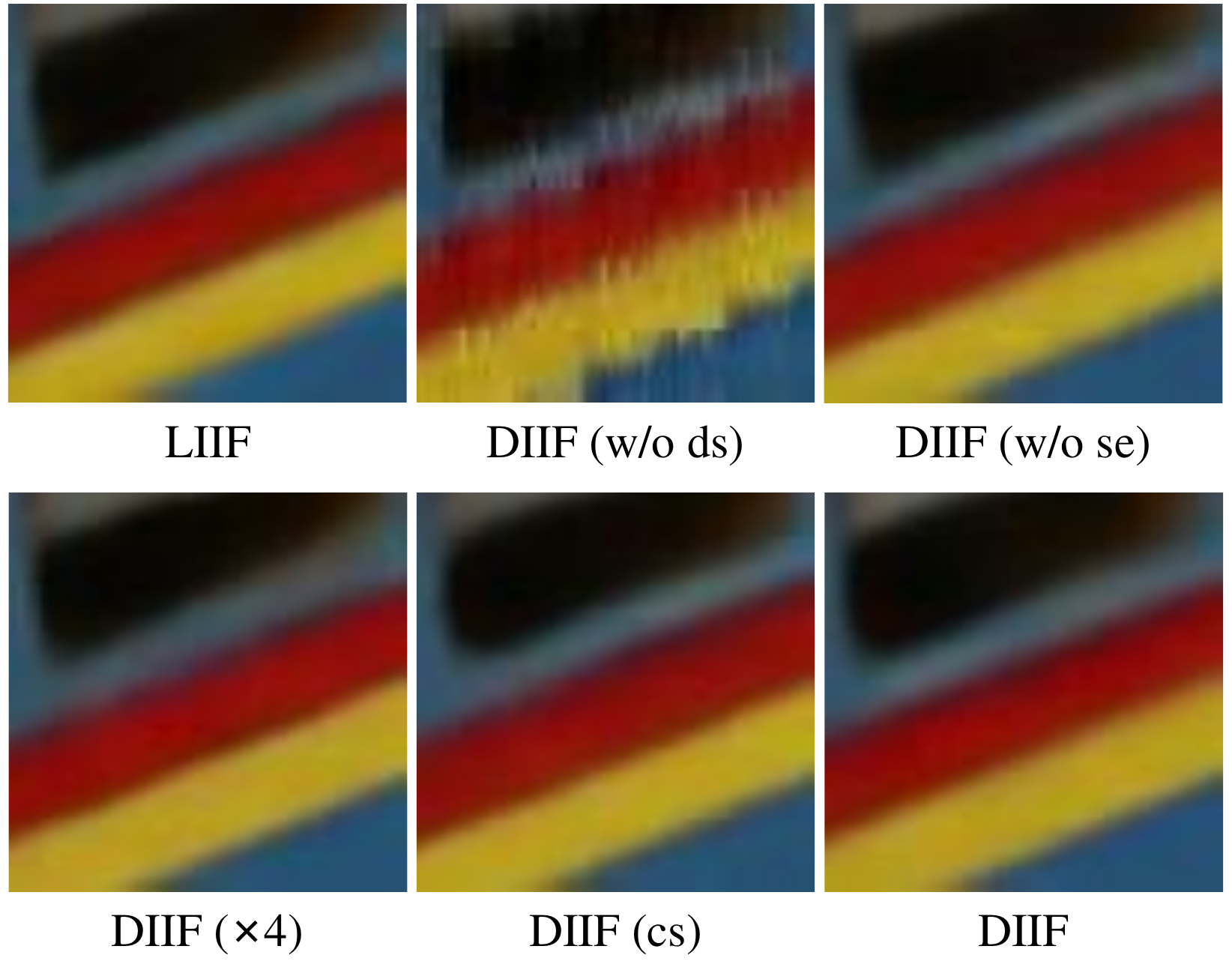}
  \caption{Qualitative results of the models in Table \ref{tab:ablation_study} on the $\times 12$ SR task. All of the models use EDSR-baseline as the encoder.}
  \label{fig:ablation_study_figure}
\end{figure}

In Table \ref{tab:ablation_study}, we conduct the ablation study on the design choices of DIIF. We provide the PSNR results on DIV2K validation set for performance evaluation, and the MACs results on the $\times16$ SR task for efficiency evaluation. In Figure \ref{fig:ablation_study_figure}, we provide the qualitative results of the models in Table \ref{tab:ablation_study} on the $\times 12$ SR task. 

\noindent \textbf{Coordinate Slicing Strategy.}
In Table \ref{tab:ablation_study}, we evaluate using different coordinate slicing strategies. DIIF (w/o ds) uses fixed coordinate slicing, which sets a fixed slice interval $4$ for any scale. Both DIIF and DIIF (cs) use dynamic coordinate slicing, which adjusts the slice interval as the scale factor varies. The results in Table \ref{tab:ablation_study} show that DIIF (w/o ds) provides poor SR performance for out-of-distribution scales compared to DIIF and DIIF (cs). Hence, it confirms that dynamic coordinate slicing is essential to the generalization to out-of-distribution scales. Another advantage of dynamic coordinate slicing is that it uses less computational cost than fixed coordinate slicing when the scale factor is large.

DIIF uses the linear-order slicing strategy, while DIIF (cs) uses the constant-order slicing strategy. DIIF achieves better PSNR performance than DIIF (cs) for all scales. The improvement in PSNR indicates the effectiveness of linear-order slicing, which divides a coordinate group into more slices for more precise prediction. Moreover, as shown in Figure \ref{fig:ablation_study_figure}, the $\times 12$ SR result of DIIF is more continuous than that of DIIF (cs). Despite the higher computational cost, we choose to use linear-order slicing in our final DIIF model.

\noindent \textbf{Slice Ensemble.}
To confirm the benefits of using slice ensemble, we compare DIIF and DIIF (w/o se) in Table \ref{tab:ablation_study}. The improvement in PSNR results demonstrates that slice ensemble improves the generalization to large scales as we designed. Moreover, as shown in Figure \ref{fig:ablation_study_figure}, the $\times 12$ SR result of DIIF achieves more continuous prediction than that of DIIF (w/o se). Despite the increase in computational cost, slice ensemble is crucial for generating continuous images with high fidelity when encountering large scales. Therefore, we use slice ensemble in our final DIIF model.

\noindent \textbf{Training with a Specific Scale.}
DIIF is trained with uniformly sampled scale factors in the range of $[\times2, \times4]$. To evaluate the performance of DIIF when it is trained with a specific scale, we train DIIF ($\times 4$) with a fixed scale factor $\times 4$. In Table \ref{tab:ablation_study}, we observe that training with fixed scale factor $\times 4$ damages the PSNR performance for all scales compared with DIIF. However, given the limited image data encountered during training, the worse performance of DIIF ($\times 4$) is to be expected. In conclusion, training with multiple scale factors can help DIIF generalize to out-of-distribution scales, which is applied to our final DIIF model.

\noindent \textbf{Coarse-to-Fine MLP.}
In Table \ref{tab:ablation_study}, we evaluate the effects of different numbers of hidden layers in C2F-MLP. Specifically, we consider the coarse stage with 2 or 4 layers and the fine stage with 3 or 5 layers. As reported in the table, a deep fine stage can decode the coarse state output better by sacrificing computational efficiency, thus providing better SR performance. However, a shallow coarse stage not only achieves competitive SR performance but also brings efficiency advantages. Overall, the performance gain from increasing the number of layers is minimal, but the computational cost increases significantly. Therefore, C2f-MLP with a shallow coarse stage (2 layers) and a shallow fine stage (3 layers) is adopted in our final DIIF model. 


\section{Conclusion}
In this paper, we propose Dynamic Implicit Image Function (DIIF) for fast and efficient arbitrary-scale image representation. In DIIF, a pixel-based image is represented as a 2D feature map and a decoding function that takes a coordinate slice and its neighboring features as inputs to predict the corresponding pixel values. By sharing the neighboring features within a coordinate slice, DIIF can be extended to large-scale SR for practical use with limited computational cost. Experimental results demonstrate that DIIF can be integrated with implicit arbitrary-scale SR methods and improve their SR performance with significantly superior computational efficiency. Better image representations and more efficient architectures for the decoding function will be explored in future work.

{\small
\bibliographystyle{ieee_fullname}
\bibliography{egbib}
}

\end{document}